 \documentclass[final,3p,times,twocolumn]{elsarticle}

\usepackage{amssymb}
\usepackage{graphicx}
\usepackage{csvsimple}
\usepackage{booktabs}
\usepackage{arydshln}
\usepackage{algorithmic,algorithm}
\usepackage{amsmath}
\usepackage{breqn}
\usepackage{lipsum}
\usepackage{adjustbox}

\journal{Sustainable Computing: Informatics and Systems (SUSCOM)}

\begin{document}

\begin{frontmatter}



\title{Optimizing a domestic battery and solar photovoltaic system with deep reinforcement learning}

\author[1]{Alexander J. M. Kell}
\author[2]{A. Stephen McGough}
\author[2]{Matthew Forshaw}

\address[1]{Sustainable Gas Institute, Imperial College London, London, UK}
\address[2]{School of Computing, Newcastle University, Newcastle-upon-Tyne, UK}

%
%
%
%


\begin{abstract}
A lowering in the cost of batteries and solar PV systems has led to a high uptake of solar battery home systems. In this work, we use the deep deterministic policy gradient algorithm to optimise the charging and discharging behaviour of a battery within such a system. Our approach outputs a continuous action space when it charges and discharges the battery, and can function well in a stochastic environment. We show good performance of this algorithm by lowering the expenditure of a single household on electricity to almost \$1AUD for large batteries across selected weeks within a year.

\end{abstract}



\begin{keyword}
Battery control \sep reinforcement learning \sep optimization \sep neural networks \sep energy

\end{keyword}

\end{frontmatter}


\section{Introduction}

The lowering cost of solar photovoltaics (PV) has led to an increase in the global uptake of residential solar PV systems since 2017 \cite{iea_2020}. Renewable energy, such as solar PV, provides a low-carbon method of generating electricity. This renewable energy can be used to decarbonise multiple sectors to help avoid the most catastrophic effects of climate change \cite{Masson-Delmotte2018}.

Australia has one of the highest rates of residential solar adoption, where 20\% of households contain solar panels~\cite{Zander2019}. However, such a high penetration of intermittent renewable energy sources can lead to issues when matching electricity supply with electricity demand, which must equal at all times. High penetrations of solar PV can be particularly challenging to manage due to the occurrence of the duck curve. The duck curve is the effect that a large amount of solar PV can have on the electricity grid, where electricity load during the day is met by solar PV. However, this output in PV is significantly reduced after nightfall \cite{kosowatz2018energy,Hou2019}.

\begin{figure}[hbt]
  \includegraphics[width=0.49\textwidth]{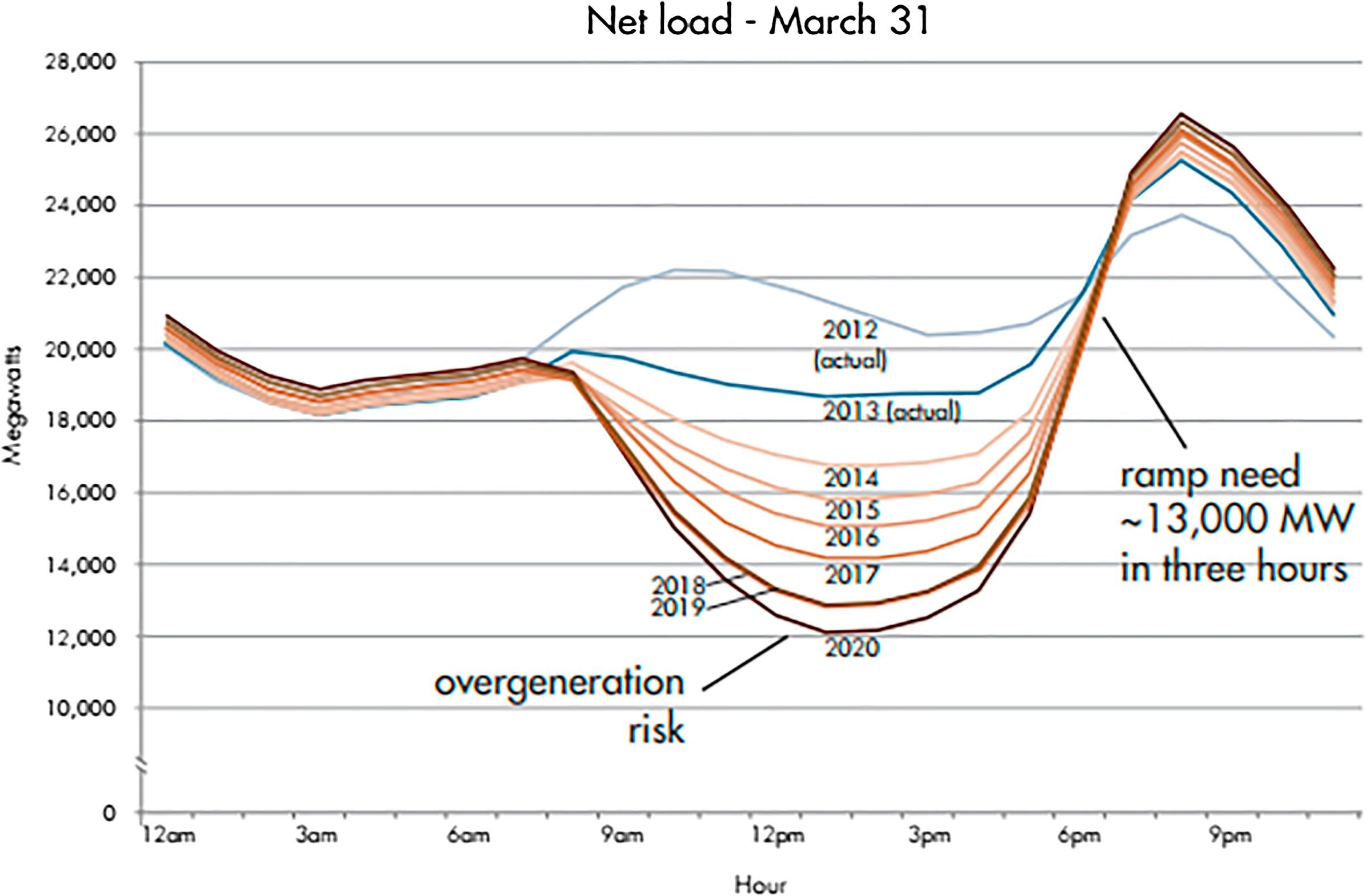}
  \caption{Duck curve for California from 2012 to 2020 \cite{casio}}
  \label{fig:duck_curve}
\end{figure}

Figure \ref{fig:duck_curve} displays an example duck curve for California from 2012 to 2020. As can be seen, as the supply of solar PV increases, a higher ramping of electricity production from non-PV sources is required after ${\sim}$5pm. This ramping rate requires dispatchable power to reduce and increase electricity supply at the required time. Dispatchable power is an electricity supply that can be used at the will of the operator. Examples include gas and coal power plants.

Batteries could provide the ability to flatten this curve by storing energy in times of high supply and low demand and discharging in times of low supply and high demand, such as in the evening. Batteries also have the ability to reduce the carbon intensity of the electricity supply. For example, instead of relying on carbon-intensive dispatchable energy sources to provide flexibility to the grid, batteries charged with low-carbon solar can be used.

The additional flexibility of the battery can be particularly useful for residential users, who would like to minimise their reliance on the grid, reduce carbon emissions for energy use, and take advantage of energy arbitrage. Some households, therefore, have begun to invest in residential battery sources. In Australia, for example, it is estimated that around 15\% of new PV installations now include battery energy storage \cite{clean_energy_council_2018}.

However, the optimal charging and discharging of a battery system is highly uncertain and stochastic. This is because there are several unknowns such as electricity demand by tariff and solar power over the future time horizon. Therefore, a system which could be used to forecast future electricity supply and demand, as well as charge and discharge the battery, would be desirable.

For this aim, this work proposes using a deep reinforcement learning (RL) algorithm to control the charging and discharging of homes with solar PV and a battery in New South Wales, Australia. We accessed a dataset provided by Ausgrid which contains 3.5 years worth of data for 300 homes, between 1 July 2010 to 30 June 2013 \cite{ausgrid}.

\begin{figure}[hbt]
  \includegraphics[width=0.49\textwidth]{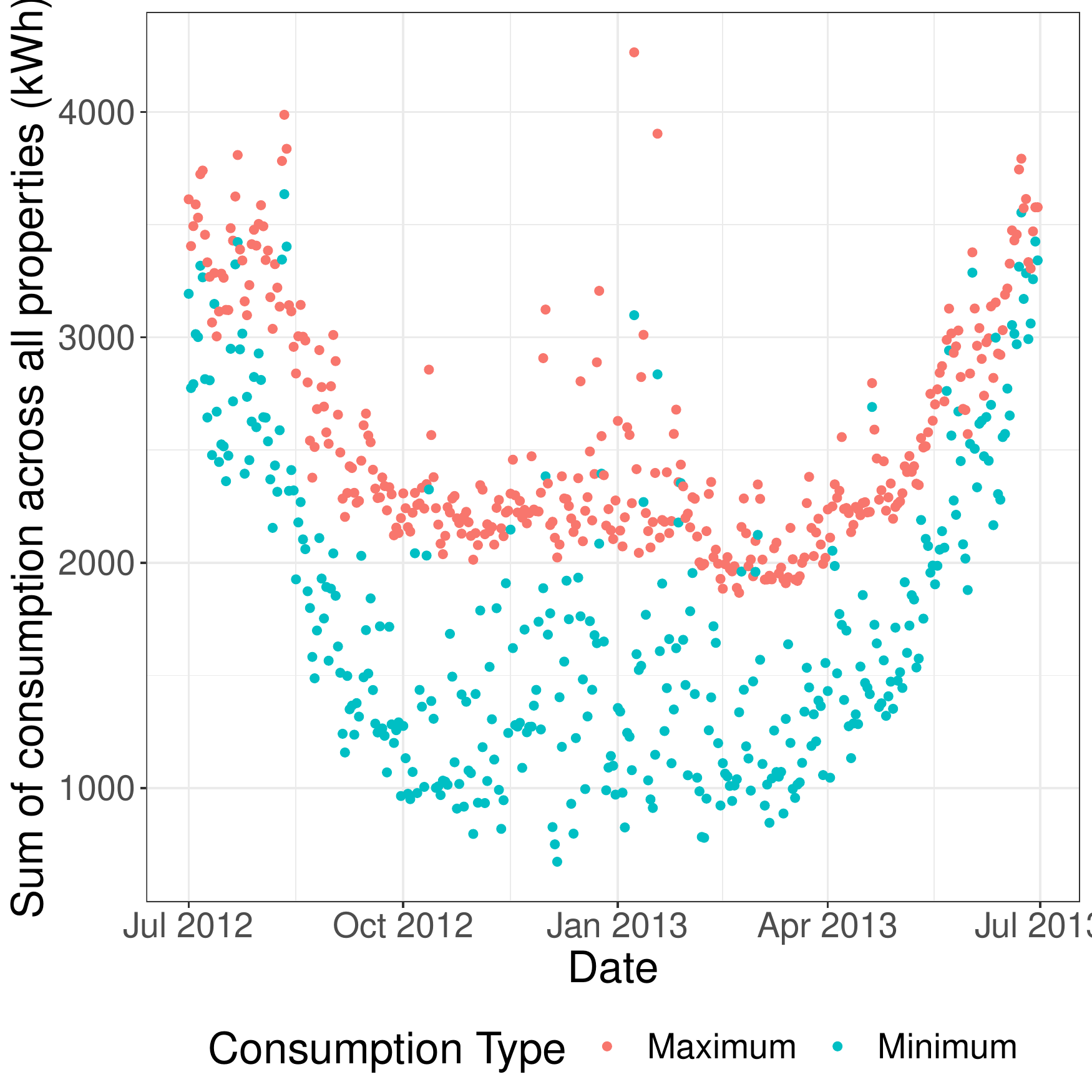}
  \caption{Maximum and theoretical minimum electricity consumption for all households.}
  \label{fig:intro-fig}
\end{figure}

Figure \ref{fig:intro-fig} displays an exploratory data analysis of the Ausgrid dataset. It demonstrates the minimum and maximum electricity consumption between July 2012 and July 2013. The red points show the load without the impact from solar power, whereas the blue points shows the theoretical minimum that the load can be reduced by using solar power. As can be seen, the load is reduced significantly by using all of the solar power. However, a simple rule can not be used to maximise the utility of the solar power. The aim of this paper is to minimise net spend on electricity consumption through the use of a battery, to get closer to the theoretical minimum (blue points).

We use the Deep Deterministic Policy Gradient (DDPG) \cite{Hunt2016a} algorithm since it uses a continuous action space and neural networks. A deep neural network is an artificial neural network with multiple layers between the input and output layers \cite{bengio2009learning}. This enables them to model complex non-linear relationships, such as those observed in electricity demand and solar PV output. 

A continuous action space allows for more precise actions that the battery makes to charge or discharge. In this work, the battery is able to precisely control the amount of charge from solar or from the grid, as well as the discharge size. A discrete action space would require us to split the action space into an integer number of actions. This leads to high computational complexity by increasing the number of action spaces to gain high precision, or losing precision by selecting too few actions. Finding the optimal number of discrete action spaces would require further study. We chose to use the DDPG algorithm due to recent success in its use in other studies \cite{Ye2020a, Kell2020d}. However, it is possible that other algorithms could also be used with success. Although, trialling all possible RL algorithms is outside of the scope of this study.

In this work, we trial ten different battery sizes (between 0.2kWh and 2kWh in 0.2 step intervals) to investigate the impact of battery size on electricity cost using data for a single household over a single year period. We split the data into a testing and training set to observe how our algorithm would function in a real-world scenario. We then trial multiple hyperparameters of a single battery to investigate whether we can achieve a higher reward.

The main contributions of this work are threefold:
\begin{enumerate}
	\item Use of a deep reinforcement learning algorithm to optimise a household solar PV battery system in Australia with a continuous action space.
	\item Investigation of the optimal battery size to reduce the usage of electricity from the grid and electricity cost.
	\item Hyperparameter tuning to find an optimal algorithm to reduce expenditure.
\end{enumerate}

We survey the literature in Section \ref{sec:lit-review} and introduce the DDPG algorithm within Section \ref{sec:material}. We present our simulation and formulate our problem in Section \ref{sec:model}. We present our results in Section \ref{sec:results} and conclude in Section \ref{sec:conclusion}.

\section{Literature Review}
\label{sec:lit-review}

The scheduling and optimisation of battery behaviour have garnered much attention. The literature proposes a wide range of approaches, spanning optimisation and artificial intelligence. In this section we review, and situate our work in the literature.

\subsection{Optimisation methods}

One approach to solving the optimal scheduling of energy storage problem is to set it up as a standard optimisation problem \cite{DeHoog2018}. Optimisation problems can be solved using linear programming, quadratic programming or mixed-integer linear programming (MILP). However, these methods require perfect knowledge of the system over the future horizon, so they cannot account for the uncertainty of the future solar supply, or electricity demand.  

Another common approach is to calculate the charging schedule using stochastic dynamic programming. These methods allow non-linear models to be incorporated into the dynamic program's cost function \cite{DeHoog2018}. However, dynamic programming models require a discretisation of the action space. This can lead to an exponential state space, and thus lead to long solution times. 

Hoke \textit{et al.} \cite{Hoke2013} apply linear programming to the economic dispatch of grid connected microgrids. They minimise the cost to operate the microgrid while meeting various constraints. They find they are able to quickly and reliably compute optimal scheduling. However, they rely on perfect foresight.

Hannah \textit{et al.} \cite{Hannah2011} propose a novel method to solve stochastic storage problems, that uses mathematical programming. They cluster states with a Dirichlet process mixture model and then fit a shape-restricted value function within each cluster. However, changes in the training data can produce large changes in policy. We overcome this issue by using neural network based RL and testing the algorithm on unseen test data.

Keerthisinghe \textit{et al.} \cite{Keerthisinghe2016} propose an approximate dynamic programming approach with temporal difference learning for implementing a home energy management system. They find that they can speed up computation compared to mixed-integer linear optimisation and dynamic programming. 

Our work's approach relaxes both the perfect knowledge of the future and discretised state space constraints. Our approach can make predictions of the future in terms of the stochastic solar power and electricity demand of a particular house.

\subsection{Deep reinforcement learning}

Deep reinforcement learning (DRL) has been highly utilised to address multiple problems, such as automated bidding \cite{Kella} and improving ride-hailing performance of electric vehicles \cite{Pettit2019a}. DRL algorithms are good approaches to tackle problems with multiple uncertainties and long time horizons, such as the problem posited in this paper. In addition, DRL agents are adaptive to test data, without the requirement to retrain the model \cite{Huang2020}.

Cao \textit{et al.} propose a charging and discharging strategy for energy storage using  Dueling Deep Q Networks (DDQN) \cite{Cao2020}. They consider price uncertainty and battery degradation and show improved effectiveness compared with a model based on mixed-integer linear programming. However, the DDQN requires a discrete action space, making finding a precise optima difficult and computationally expensive. This contrasts to our work which uses a continuous action space.

Wang \textit{et al.} \cite{Wang2018a} derive an arbitrage policy for energy storage operation in real-time markets using Q-learning to control the charging and discharging of energy storage. They show that by incorporating information about the history, they can significantly improved performance. The Q-learning algorithm, however, also uses a discrete action space.

Finally, Huang \textit{et al.} use the DDPG algorithm to solve the capacity scheduling problem. Using the DDPG algorithm, like us, can output a continuous action space, which reduces computational complexity.  However, their work tests their results on only one season, whereas we test on a randomly sampled number of weeks throughout the year.

\section{Deep Reinforcement Learning}
\label{sec:material}

Here we describe the RL methodology used for the intelligent bidding process, and the simulation model used as the environment.

\subsection{Reinforcement Learning background}

In reinforcement learning (RL) an agent interacts with an environment to maximize the cumulative reward gained from an environment by making certain actions. RL can be described as a Markov Decision Process (MDP). An MDP includes a state-space $\mathcal{S}$, action space $\mathcal{A}$, a transition dynamics distribution $p(s_{t+1}|s_t,a_t)$ and a reward function, where $r:S\times \mathcal{A} \rightarrow \mathbb{R}$. The agent's behaviour is modified by an observation of the current state in each time step. In other words, the agent adjusts its actions based on its previous actions and observations that arise from its actions.

An agent's behaviour is defined by a policy, $\pi$. $\pi$ maps states to a probability distribution over the actions $\pi:\mathcal{S}\rightarrow \mathcal{P}(\mathcal{A})$. The return from a state is defined as the sum of discounted future reward $R_t=\sum_{i=t}^T\gamma^{(i-t)}r(s_i,a_i)$. Where $\gamma$ is a discounting factor and $\gamma \in [0,1]$. The reward is dependent on the action chosen, which is specified by policy $\pi$. The goal of reinforcement learning is to learn a policy that maximizes the expected cumulative return from the start distribution $J=\mathbb{E}_{r_i,s_i \sim E,a_i \sim \pi}[R_1]$. Or, in other words, to gain the highest cumulative reward from the environment.

The expected return after taking an action $a_t$ in state $s_t$ after following policy $\pi$ can be found by the action-value function. The action-value function is used in many reinforcement learning algorithms and is explicitly defined in Equation \ref{eq:action-value}.
\begin{equation}
	\label{eq:action-value}
	Q^{\pi}(s_t,a_t)=\mathbb{E}_{r_{i\geq t},s_{i>t}\sim \mathcal{E},a_{i>t}\sim\pi}[R_t|s_t,a_t].
\end{equation}
\noindent The action-value function defines the expected reward at time $t$, given a state $s_t$ and action $a_t$ when under policy $\pi$. It is effectively a function which predicts the reward from a certain action in the environment. A goal of the RL algorithm is to learn such a function so that the correct action is taken to maxmimize the reward.

\subsection{Q-Learning}

 An optimal policy can be derived from the optimal $Q$-values 
 
\begin{equation}
Q_*(s_t,a_t)=\max_\pi Q_\pi(s_t,a_t). 
\end{equation} 
 
 \noindent Q-Learning works by selecting the action corresponding to the highest Q-value in each state.

Many approaches in reinforcement learning use the recursive relationship known as the Bellman equation, as defined in Equation \ref{eq:bellman}:
\begin{dmath}
	\label{eq:bellman}
	Q^\pi(s_t,a_t)=\mathbb{E}_{{r_t},s_{t+1}\sim E} [r(s_t,a_t)+
	\gamma\mathbb{E}_{a_{t+1}\sim \pi}[Q_\pi(s_{t+1},\pi(s_{t+1}))]].
\end{dmath}
\noindent The Bellman equation is equal to the action which maximizes the reward plus the discount factor multiplied by the next state's value, which would occur after following the policy in state $s_{t+1}$ or $\pi(s_{t+1})$.

The Q-value can therefore be improved by bootstrapping. Bootstrapping is where the current value of the estimate of $Q_\pi$ is used to improve its estimate of the future, using the known $r(s_t,a_t)$ value. This is the foundation of Q-learning \cite{Gay2007}, a form of \textit{temporal difference} (TD) learning \cite{Sutton2015}, where the update of the Q-value after taking action $a_t$ in state $s_t$ and observing reward $r_t$, which results in state $s_{t+1}$ is:
\begin{equation}
	Q(s_t,a_t)\leftarrow Q(s_t,a_t)+\alpha\delta_t,
\end{equation}
\noindent where,
\begin{equation}
	\delta_t=r_t+\gamma\max_{a_{t+1}}Q(s_{t+1},a_{t+1})-Q(s_{t},a_t),
\end{equation}
\noindent $\alpha\in [0,1]$ is the step size, $\delta_t$ represents the correction for the estimation of the Q-value function and $r_t+\gamma\max_{a_{t+1}}Q(s_{t+1},a_{t+1})$ represents the target Q-value at time step $t$.

It has been proven that if the Q-value for each state action pair is visited infinitely often, the learning rate $\alpha$ decreases over time step $t$. So, as $t\rightarrow \infty$, $Q(s,a)$ converges to the optimal $Q_*(s,a)$ for every state-action pair \cite{Gay2007}. However, Q-learning often suffers from the curse of dimensionality, because the Q-value function is stored in a look-up table which therefore requires the action and state spaces to be discretized. As the number of discretized states and actions increases, the computational cost increases exponentially, making the problem intractable. 

\subsection{Deep Deterministic Gradient Policy}

Many problems are naturally discretized which are well suited to a Q-learning approach, however this is not always the case, such as the battery control problem we investigate here. It is not, however, straightforward to apply Q-learning to continuous action spaces. This is because in continuous spaces, finding the greedy policy requires an optimization of $a_t$ at every time step. Optimizing for $a_t$ at every time step would be too slow to be practical with large, unconstrained function approximators and nontrivial action spaces \cite{Hunt2016a}. To solve this, an actor-critic approach based on the deterministic policy gradient (DPG) algorithm is used \cite{Silver2014}.

The DPG algorithm maintains a parameterized actor function $\mu(s|\theta^\mu)$ which specifies the current policy by deterministically mapping states to a specific action. The critic $Q(s,a)$ is learned using the Bellman equation, as in Q-learning. The actor is updated by applying the chain rule to the expected return from the start distribution $J$ with respect to the actor parameters:
\begin{align}
\begin{split}
	\triangledown_{\theta^\mu}J\approx\mathbb{E}_{s_t\sim\rho^\beta}[\triangledown_{\theta^\mu}Q(s,a|\theta^Q)|_{s=s_t,a=\mu(s_t|\theta^\mu)}] \\
	= \mathbb{E}_{s_t\sim\rho^\beta}[\triangledown_aQ(s,a|\theta^Q)|_{s=s_t,a=\mu(s_t)}\triangledown_{\theta_\mu}\mu(s|\theta^\mu)|_{s=s_t}].
\end{split}
\end{align}
 
This is the policy gradient, the gradient of the policy's performance. The policy gradient method optimizes the policy directly by updating the weights, $\theta$, in such a way that an optimal policy is found within finite time. This is achieved by performing gradient ascent on the policy and its parameters $\pi^\theta$.

Introducing non-linear function approximators, however, means that convergence is no longer guaranteed. Although these function approximators are required in order to learn and generalize on large state spaces. The Deep Deterministic Gradient Policy (DDPG) modifies the DPG algorithm by using neural network function approximators to learn large state and action spaces online.

A replay buffer is used in the DDPG algorithm to address the issue of ensuring that samples are independently and identically distributed. The replay buffer is a finite-sized cache, $\mathcal{R}$. Transitions are sampled from the environment through the use of the exploration policy, and the tuple $(s_t,a_t,r_t,s_{t+1})$ is stored within this buffer. $\mathcal{R}$ discards older experiences as the replay buffer becomes full. The actor and critic are trained by sampling from $\mathcal{R}$ uniformly. 

A copy is made of the actor and critic networks, $Q'(s,a|\theta^{Q'})$ and $\mu'(s|\theta^{\mu'})$ respectively. These are used for calculating the target values. To ensure stability, the weights of these target networks are updated by slowly tracking the learned networks. Pseudo-code of the DDPG algorithm is presented in Algorithm \ref{alg:ddpg}.

\begin{algorithm}
\caption{DDPG Algorithm \cite{Hunt2016a}}
\begin{algorithmic}[1]
  \STATE Initialize critic network $Q(s,a|\theta^Q)$ and actor $\mu(s|\theta^\mu)$ with random weights $\theta^Q$ and $\theta^\mu$
  \STATE Initialize target network $Q'$ and $\mu'$ with weights $\theta^{Q'}\leftarrow\theta^Q,\theta^{\mu'}\leftarrow \theta^{\mu}$
  \STATE Initialize replay buffer $R$
  \FOR{\texttt{episode=1,M}}
        \STATE Initialize a random process $\mathcal{N}$ for action exploration
        \STATE Receive initial observation state $s_1$
        \FOR{\texttt{t=1,T}}
        	\STATE Select action $a_t=\mu(s_t|\theta^{\mu})+\mathcal{N}_t$ according to the policy and exploration noise, $\mathcal{N}_t$
        	\STATE Execute action $a_t$ and observe reward $r_t$ and new state $s_{t+1}$
        	\STATE Store transition $(s_t, a_t, r_t, s_{t+1})$ in $R$
        	\STATE Sample a random minibatch of $N$ transitions $(s_i, a_i, r_i, s_{i+1})$ from $R$
        	\STATE Set $y_i=r_i+\gamma Q'(s_{i+1},\mu'(s_{i+1},\mu'(s_{i+1}|\theta^{\mu'})|\theta^{Q'})$
        	\STATE Update critic by minimizing the loss: $$L=\frac{1}{N}\sum_i(y_i-Q(s_i,a_i|\theta^Q))^2$$
        	\STATE Update the actor policy using the sampled policy gradient: $$\triangledown_{\theta^\mu}J\approx \frac{1}{N}\sum_i\triangledown_a Q(s,a|\theta^Q)|_{s=s_i,a=\mu(s_i)}\triangledown_{\theta^\mu}\mu(s|\theta^\mu)|_{s_i}$$
        	\STATE Update the target networks:
        	$$\theta^{Q'}\leftarrow\tau\theta^Q+(1-\tau)\theta^{Q'}$$
        	$$\theta^{\mu'}\leftarrow\tau\theta^\mu+(1-\tau)\theta^{\mu'}$$
        \ENDFOR
      \ENDFOR
\end{algorithmic}
\label{alg:ddpg}
\end{algorithm}

Algorithm \ref{alg:ddpg} first, initialises the critic network and actor with random weights. For each episode of the simulation, the state space is received. For each time-step, an action is taken according to the policy and exploration noise. After this action a new state and reward is observed. The transition $(s_t, a_t, r_t, s_{t+1})$ is then stored in the replay buffer, $R$. A random minibatch of $N$ transitions is then taken from $R$. The critic is then updated by minimising the loss and the actor policy is updated using the sampled policy gradient. Finally, the target networks are updated. The algorithm presented here is based upon work from \cite{Hunt2016a}. The algorithm is utilised using our novel simulation.

\section{Solar-Battery Simulation}
\label{sec:model}

The simulation model developed to train the reinforcement learning algorithm is primarily based on data from Ausgrid \cite{ausgrid}. This dataset contains data regarding 300 homes with rooftop solar. A gross meter measures the total amount of power generated every 30 minutes. This data has been sourced from 300 randomly selected solar customers in Ausgrid's electricity network. 

This dataset contains solar power generated and electricity demand per tariff. The electricity demand contains two tariffs: general consumption and controlled load consumption. In Australia, the general consumption is the consumption for most appliances. Whereas the controlled load consumption is electricity supplied to specific appliances, such as electric hot water systems, slab or underfloor heating, which are often separately metered. A controlled load tariff generally has a lower rate, as these appliances operate during off-peak hours.

Using this dataset, a simulation was designed to simulate the behaviour of a solar battery system. Each half-hour from the dataset was designated as a time-step in the simulation. The action space for DDPG RL algorithm, $a_t$ is defined as follows:
\begin{equation}
a_t=(C_s, C_{cl}, D_b)	
\end{equation}

\noindent where $C_s$ is the amount of energy the battery should charge from solar power, $C_{cl}$ is the amount of charge the battery should charge from the controlled load tariff, and $D_b$ is the amount the battery discharge.

For our problem formulation, the state space, $s_t$, is given by the following tuple:
\begin{equation}
	s_t = (B_s, B_c, GC, CL, CS, R_{cl}, R_{gl})
\end{equation}

\noindent where $B_s$ is the size of the battery, $B_c$ is the charge of the battery, $GC$ is the general electricity consumption, $CL$ is the controlled load consumption, $CS$ is the solar PV power output at the current time-step, $R_{cl}$ and $R_{gl}$ are the residual controlled load electricity and residual general electricity respectively. The residual controlled load electricity and residual general electricity represent the electricity consumption that could not be met by the battery. All of these variables, apart from $B_s$ are during the time-step in question. We assume that $B_s$ is constant for the simulation. We appreciate that as the battery ages $B_s$ will reduce. However, the fundamental principle of the simulation does not change, ie. how to optimally charge and discharge a battery. In addition, we do not take into account efficiency loss between cycles, which are typically 20\%.

The reward, $r_t$, is the inverse of the total price paid for the electricity consumption which was not serviced by solar energy. This includes the controlled load, which was used to charge the battery.

Once the simulation receives the input from the RL algorithm, the battery is charged in order of priority: firstly by solar power and secondly from the controlled load tariff. We assume in this work that the controlled load tariff is from 23:00 until 8:00, as per the dataset provided by Ausgrid \cite{ausgrid}. This is due to the separate nightly tariff provided by Ausgrid at these times.

The DDPG algorithm is well suited to this action and state space as electricity is fundamentally a continuous and non-discrete quantity. The continuous action space of the DDPG algorithm can therefore more finely and precisely make actions. This is especially true for households with highly varying electricity demands, with different min-max that could be met with a battery source.

Once the battery has been charged, as per the actions of the RL algorithm, the electricity demand is serviced by the residual electricity supply, ie. the solar and controlled load which did not charge the battery. This is either due to the maximum capacity of the battery, or due to a smaller action from the RL algorithm. The remaining load is paid for at a price of AUD\$0.27/kWh for general electricity consumption and AUD\$0.10/kWh for controlled load consumption. 

Within the dataset there were multiple missing days. These were dealt with by removing all days which did not make a full week. This left us with 15 full weeks in the year 2013, which were evenly distributed amongst seasons. 15 weeks gave us enough data to adequately represent an entire year in 30-minute time-steps. This high granularity limited the total number of weeks we could run, due to the high computational time it took to run such algorithm and simulation. We model a single household for this work.

Next, we split the data into a training set and a test set, by selecting eight weeks for training and seven weeks for testing by randomly sampling. We chose this ratio to give us enough data for training, as well as to give us a broad range of weeks throughout the year in which we could test. This allowed us to test our model on a diverse dataset due to different electricity demand profiles, and solar power output from solar PV.

We limit this analysis to a single year to provide representative results for a single year. We could have run the simulation for the entire 3.5 year dataset, but this would make the results more difficult to interpret for a year time-period, as well as increasing the computational time required to gain results. 

We chose a single household to reduce training and testing time. However, we believe that this approach is adequate to maximise the reward per household. For instance, a household may have different characteristics to other households in terms of electricity demand and solar supply.


We do not take into account battery life degradation in this work, as we run the algorithm over a single year time-period. We therefore assume that the battery capacity does not degrade over this time period. As we model 30 minute time steps, we do not take into account differing discharge or charge rates, as it is assumed that battery technologies can charge and discharge equally well in 30 minute time-steps.

In addition to this, we tuned various hyperparameters for a single size of battery. For the hyperparameter training, we chose a single battery size to reduce compute time and cost. However, the same approach could be used for any battery size. For the hyperparameter tuning, we trialled the parameters as shown in Table \ref{tab:hyper-param}. The actor hiddens and critic hiddens variables is the architecture of the actor and critic networks respectively. For instance [200, 200] indicates 200 neurons in the first hidden layer and 200 neurons in the second hidden layer. The search method indicates the method in which these hyperparameters were chosen, either grid search or from a uniform distribution. A grid search means that each value is chosen with every other value, whereas the uniform distribution samples randomly between the range stated from a uniform distribution.

\begin{table}[]
\begin{adjustbox}{width=1\columnwidth}
\begin{tabular}{@{}ccl@{}}
\toprule
Variable       & Search method        & Values                                                         \\ \midrule
Actor hiddens  & Grid search          & {[}200, 200{]}, {[}300, 300{]}, {[}400, 400{]}                 \\
Critic hiddens & Grid search          & {[}200, 200{]}, {[}300, 300{]}, {[}400, 400{]}, {[}500, 500{]} \\
Learning rate  & Uniform distribution & {[}$1\times10^{-7}$, $1\times10^{-1}${]}                        \\ \bottomrule
\end{tabular}}
\end{adjustbox}
\caption{Hyperparameter variables used for tuning of the DDPG algorithm.}
\label{tab:hyper-param}
\end{table}

We used the Tune RLlib library for model training parallelisation and for the use of reinforcement learning \cite{liaw2018tune}.

\section{Results}
\label{sec:results}

In this section we investigate the results of the DDPG algorithm applied to the capacity scheduling control problem. 

\subsection{Testing and training}

\begin{figure}[]
  \includegraphics[width=0.49\textwidth]{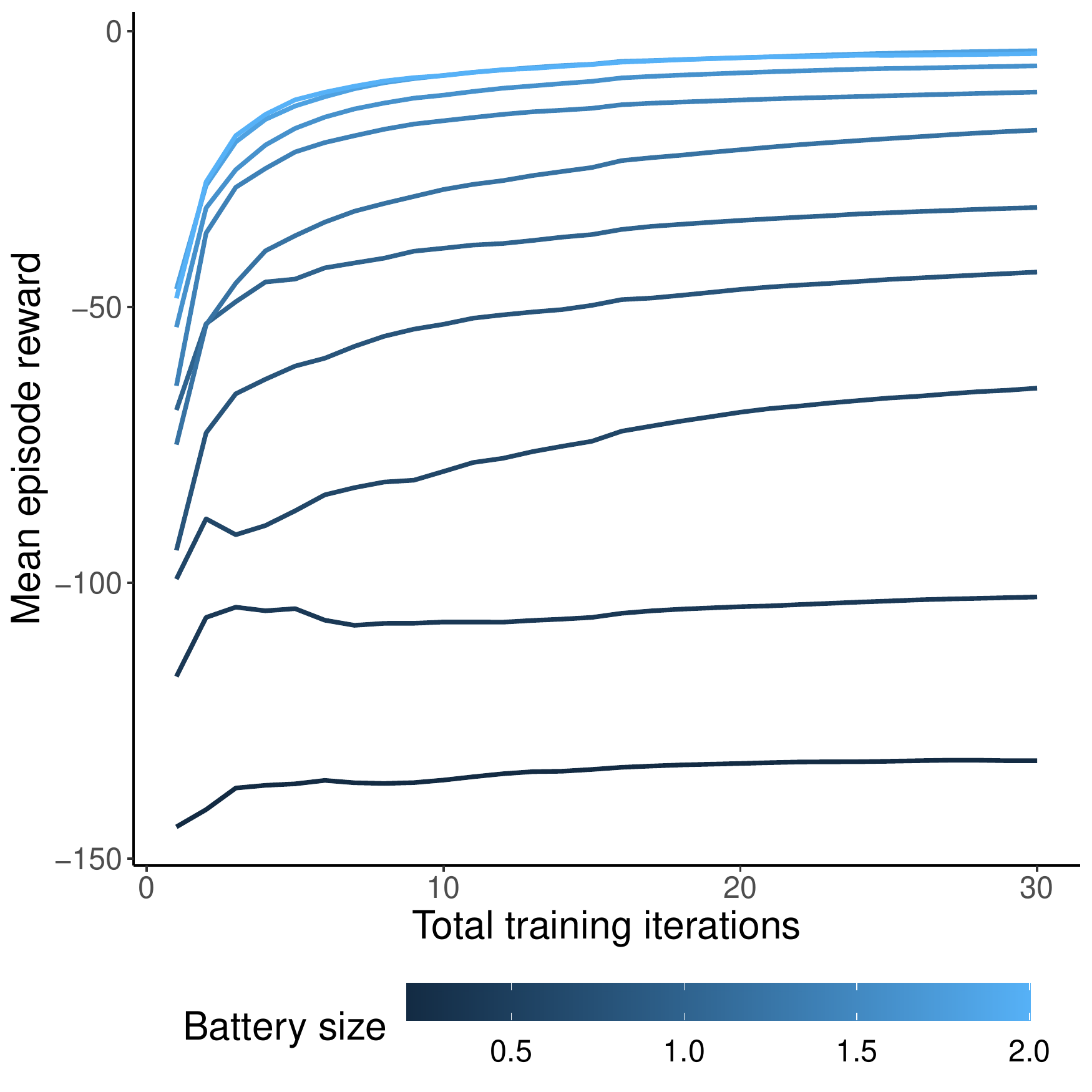}
  \caption{DDPG mean episode reward over total training iterations by battery size.}
  \label{fig:training_rl}
\end{figure}

Figure \ref{fig:training_rl} displays the total training iterations versus the mean episode reward for systems with different battery sizes. The figure shows algorithm convergence for all, but one, of the different battery sizes. That is, the reward increases over total training iterations until there are marginal gains from increasing the training time. 

Another observation is that reward increases with battery size. That is, the larger the battery, the less money is spent on purchasing electricity. This is likely because more energy can be used to service demand with solar power.

\begin{figure}[]
  \includegraphics[width=0.49\textwidth]{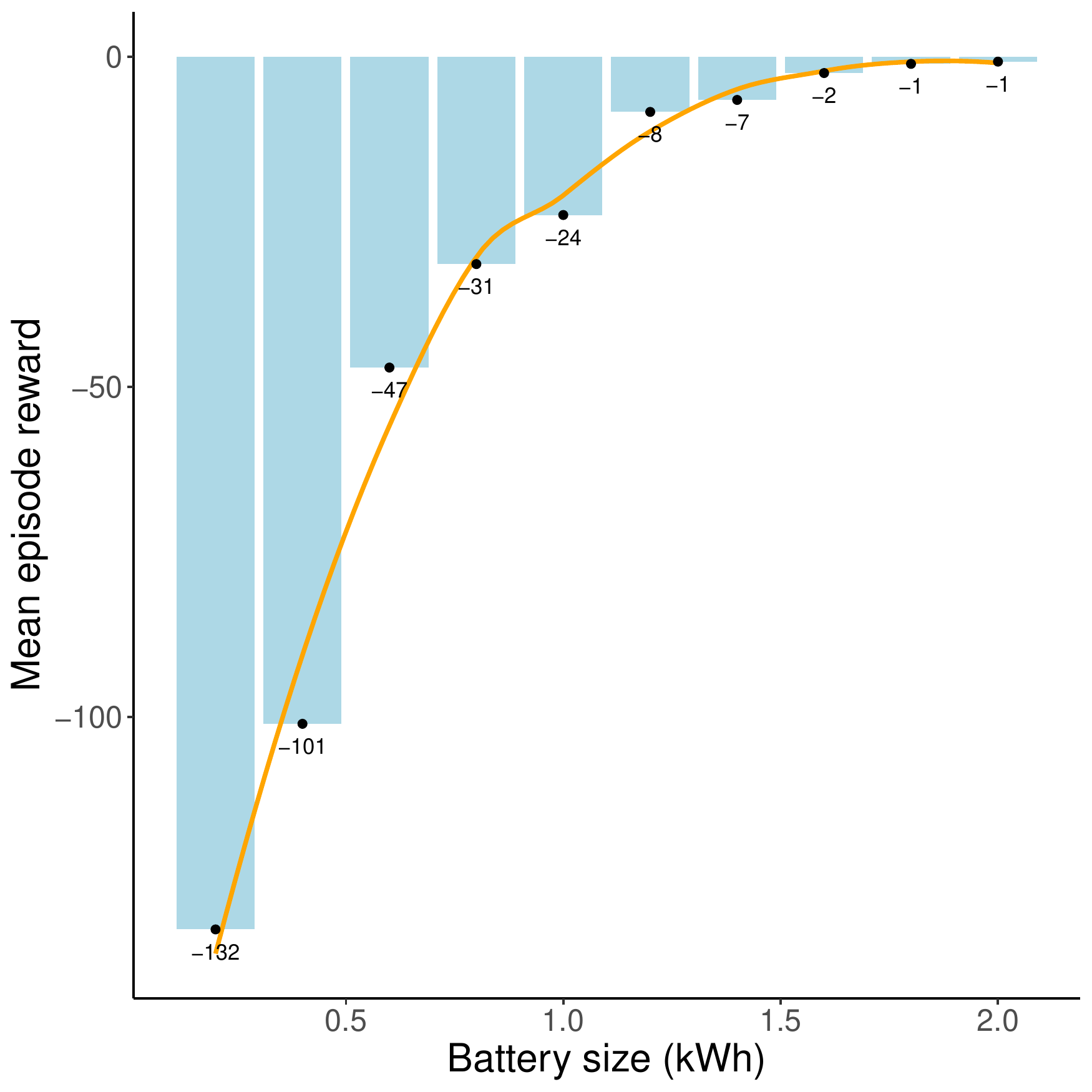}
  \caption{Testing with unseen data per battery size.}
  \label{fig:testing}
\end{figure}

Figure \ref{fig:testing} displays the mean episode reward after using the trained algorithms per battery size. As expected, the larger the battery size, the larger the reward. With increasing gains at smaller battery sizes. However, after a battery size of 1.2kWh there are diminishing returns. This suggests that it may be economically optimal to purchase a 1.2kWh battery for this particular household.

Figure \ref{fig:testing} shows that our algorithm can generalise to previously unseen data, by maintaining a high reward. With a battery of 1.8kWh and above, the mean episode reward is $-1$, which means that very little energy was bought, on average, from the grid. Specifically, AUD\$1 was spent on average on electricity per episode (8 week period). However, increasing battery capacity involves a significantly higher outlay in terms of capital expenditure. Small batteries, for instance of 0.2kWh and 0.4kWh, on the other hand significantly reduce the savings, with an outlay of AUD\$132 and AUD\$101 respectively. 

However, a decision on battery size is dependent on the preferences of the household and user. A larger house, with high electricity consumption, may benefit from a larger battery and vice-versa. In addition, certain users may prefer high capital expenditure, in return for a low operational expenditure, or vice-versa. Therefore, it is not possible to suggest the ``best'' battery size for each user.

\begin{figure}[]
  \includegraphics[width=0.49\textwidth]{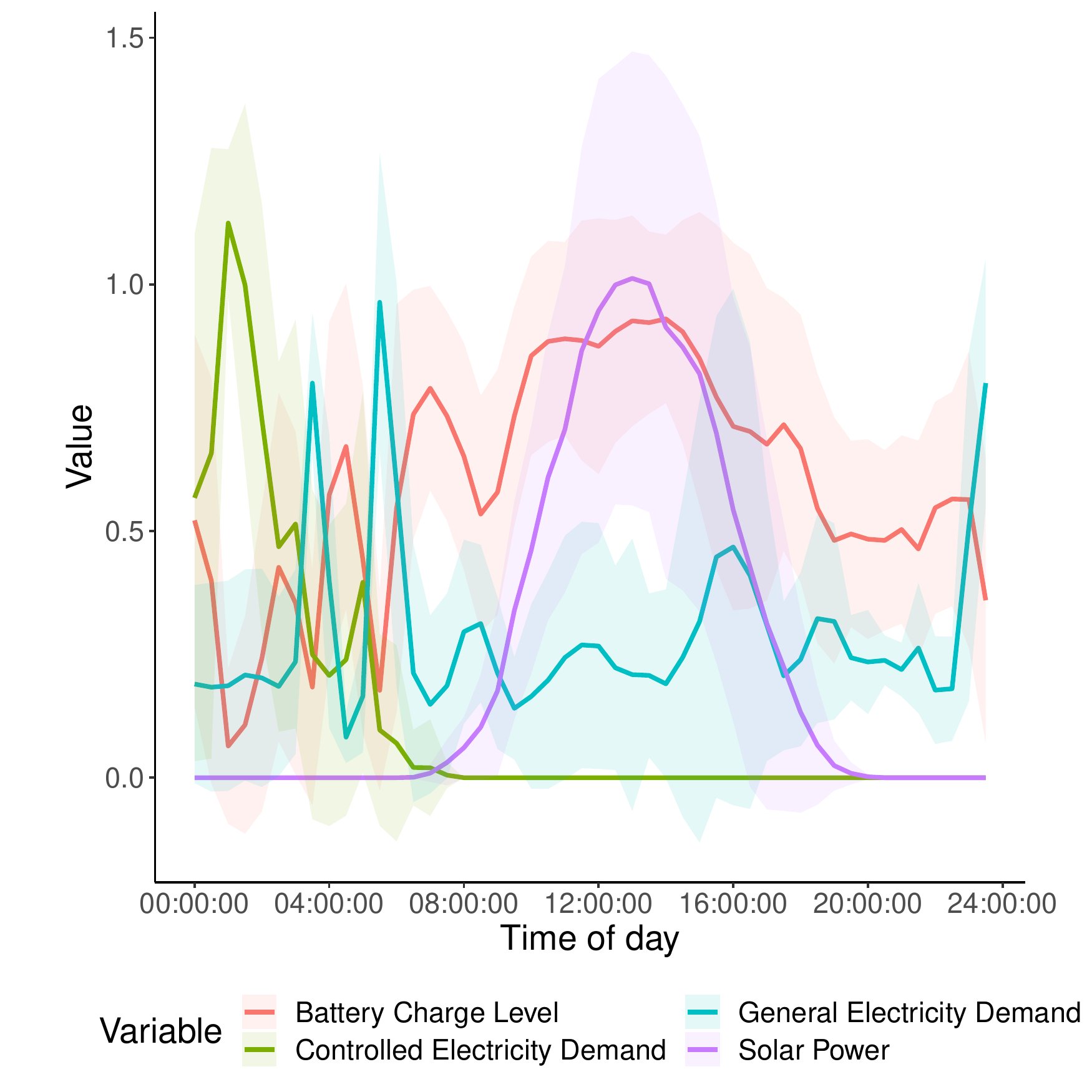}
  \caption{Battery of 1kWh controlled by reinforcement learning algorithm over a year for days with high mean episode reward ($>$-30).}
  \label{fig:average_day}
\end{figure}

Figure \ref{fig:average_day} displays the mean and standard deviation of the household with a 1kWh battery on the days with a high mean episode reward ($\textgreater$-30).

This plot shows that the DDPG algorithm chooses to charge the battery (orange line) as much as possible with solar power. This power is then immediately discharged after the sun sets to meet the general electricity demand (blue line). The battery then charges the controlled electricity demand to charge the battery and discharge the general electricity demand as much as possible. A large amount of uncertainty exists in these plots, shown by the standard deviation, due to the different demand profiles and solar irradiance. However, the battery charge profile is able to meet this uncertainty, by having a large standard deviation itself.

\subsection{Hyperparameter tuning}

As discussed in Section \ref{sec:model}, we tuned various hyperparameters to find the DDPG algorithm with the best reward. Table \ref{tab:hyper-params-result} displays the rewards of the best and worst eight parameters. We see that there is a difference of over 56\% between the best and worst hyperparameter set with the same algorithm. This displays the importance of hyperparameter tuning in this work.

\begin{table}[]
\begin{adjustbox}{width=1\columnwidth}
\begin{tabular}{@{}ccccc@{}}
\toprule
Rank & LR     & Actor hiddens  & Critic hiddens & Mean episode reward \\ \midrule
1    & 0.0297 & {[}200, 200{]} & {[}300, 300{]} & -23.6664            \\
2    & 0.0577 & {[}200, 200{]} & {[}300, 300{]} & -23.9457            \\
3    & 0.0592 & {[}400, 400{]} & {[}300, 300{]} & -24.2017            \\
4    & 0.0448 & {[}400, 400{]} & {[}200, 200{]} & -24.9747            \\
5    & 0.0941 & {[}300, 300{]} & {[}400, 400{]} & -25.3434            \\
6    & 0.0659 & {[}400, 400{]} & {[}300, 300{]} & -25.4411            \\
7    & 0.0712 & {[}300, 300{]} & {[}400, 400{]} & -25.8312            \\
8    & 0.0699 & {[}400, 400{]} & {[}500, 500{]} & -25.8656            \\
\hdashline

65   & 0.0486 & {[}200, 200{]} & {[}200, 200{]} & -33.3017            \\
66   & 0.036  & {[}200, 200{]} & {[}200, 200{]} & -33.3643            \\
67   & 0.036  & {[}300, 300{]} & {[}500, 500{]} & -34.1956            \\
68   & 0.0709 & {[}400, 400{]} & {[}200, 200{]} & -34.5363            \\
69   & 0.0216 & {[}400, 400{]} & {[}500, 500{]} & -34.8443            \\
70   & 0.0905 & {[}400, 400{]} & {[}200, 200{]} & -34.899             \\
71   & 0.0987 & {[}200, 200{]} & {[}400, 400{]} & -35.2939            \\
72   & 0.0227 & {[}200, 200{]} & {[}200, 200{]} & -36.1401            \\ \bottomrule
\end{tabular}
\end{adjustbox}
\caption{Best and worst hyperparameter combinations for the optimisation of a 1kWh solar battery system.}
\label{tab:hyper-params-result}
\end{table}

\begin{figure}[]
  \includegraphics[width=0.49\textwidth]{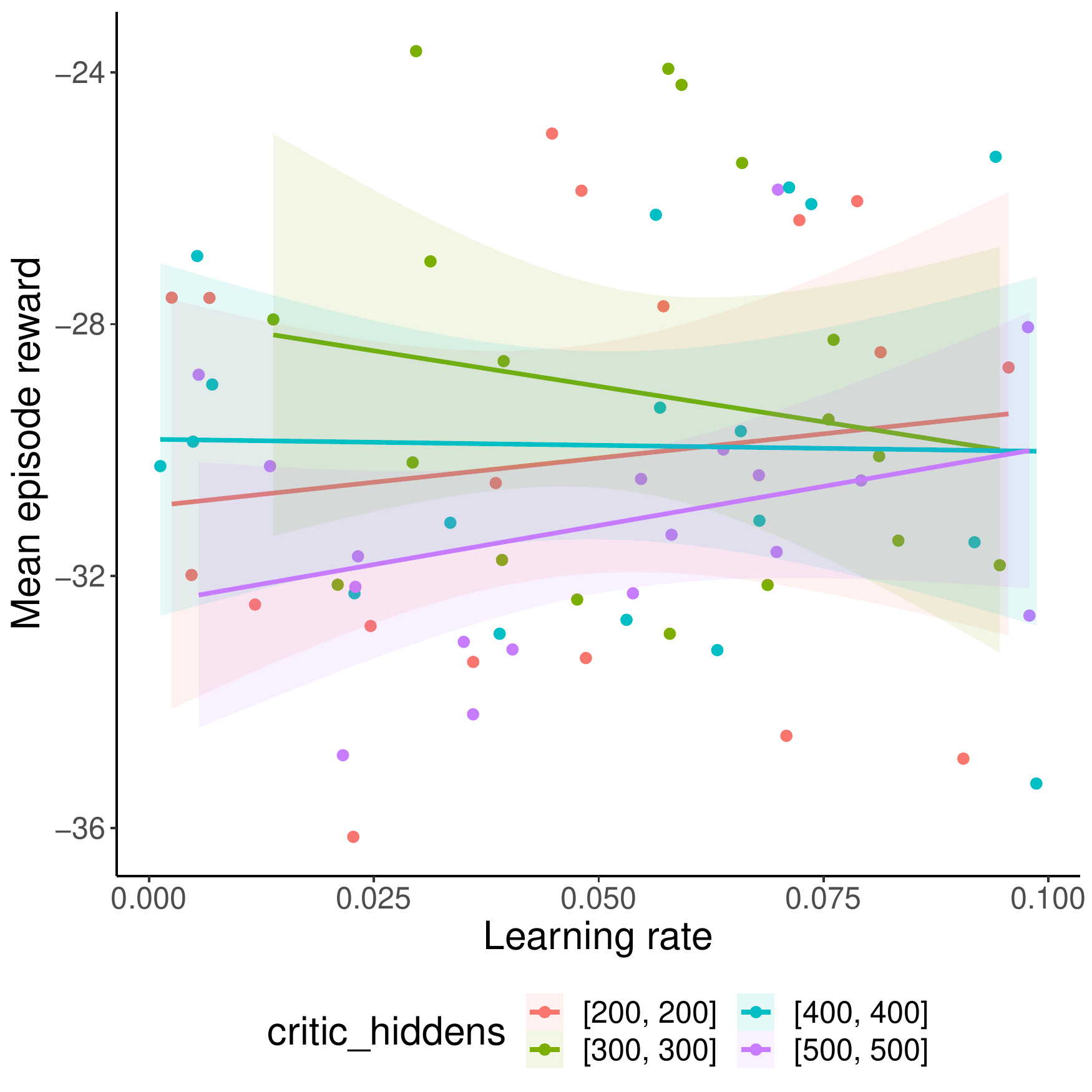}
  \caption{Hyperparameter tuning for a battery size of 1kWh.}
  \label{fig:hyperparameter-plot}
\end{figure}

Figure \ref{fig:hyperparameter-plot} shows mean episode reward versus learning rate, with the colours showing the critic hidden layer. We can see that for a layer of [300, 300] a lower learning rate is optimal, which shows the highest reward. Whereas for any of the other architectures, a higher learning rate is optimal. Albeit showing lower results.

\begin{figure}[]
  \includegraphics[width=0.49\textwidth]{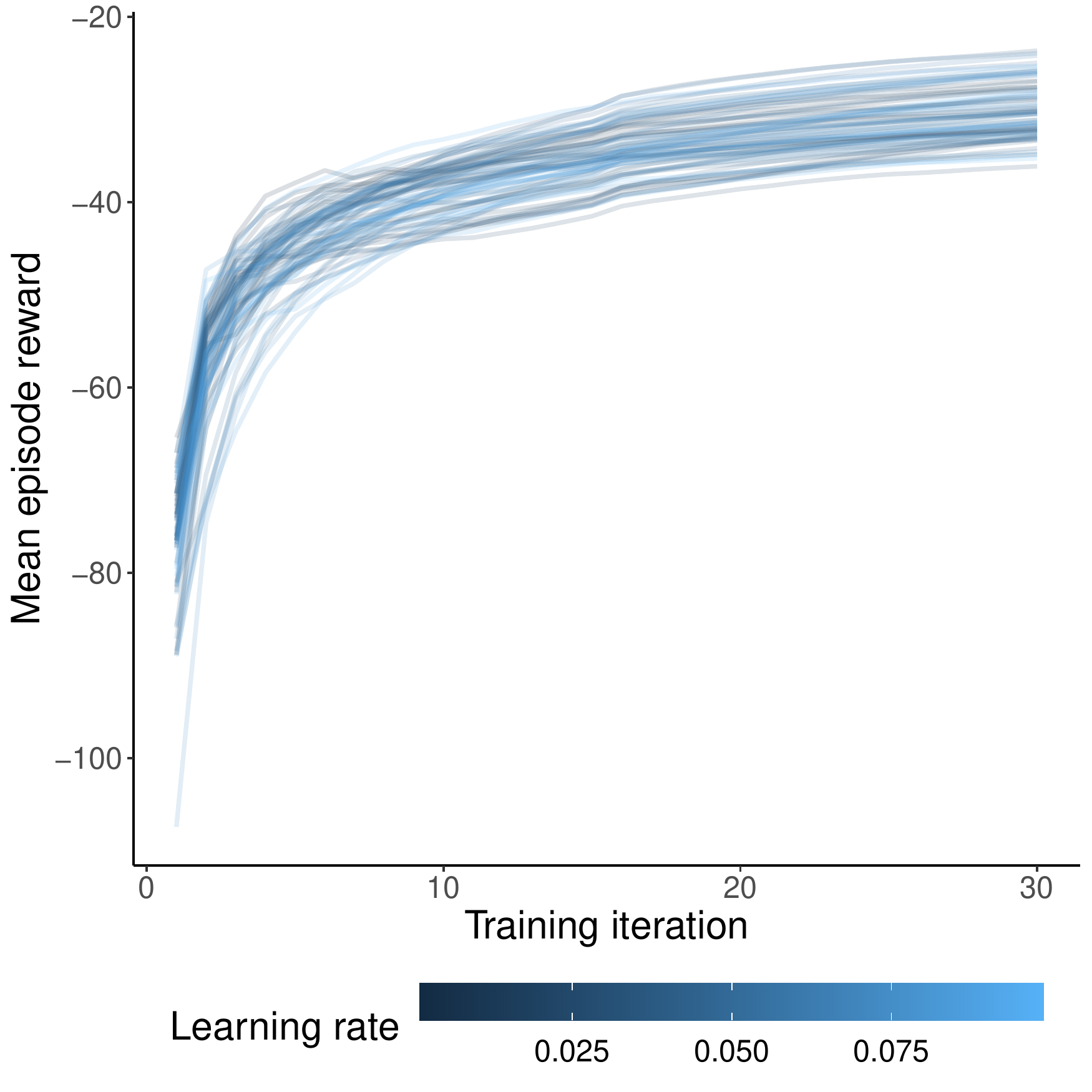}
  \caption{Hyperparameter tuning convergence for a battery size of 1kWh.}
  \label{fig:hyperparameter-training}
\end{figure}

Figure \ref{fig:hyperparameter-training} displays the hyperparameter tuning training process. Whilst all algorithms follow a similar pattern, there is a large range in rewards across all training iterations. Moreover, it can be seen that the learning rate is not linearly correlated with the reward. 

\section{Conclusion}
\label{sec:conclusion}

In this work, we modelled a solar battery home system in New South Wales, Australia. We used the deep deterministic policy gradient (DDPG) algorithm to control charging and discharging of various battery sizes. We were able to model a continuous action space, reducing the computational complexity of a high dimensional discrete action space whilst giving us more precise results. We showed that we were able to achieve good results, by both charging and discharging the battery  in a stochastic environment.
 
We tuned the hyperparameters of the model to be optimal for a 1kWh battery for a single house. We showed that we were able to achieve an improvement in 56\% over the worst parameter set through hyperparameter tuning.

In future work we would like to add a battery degradation model to the simulation to incorporate degradation of the battery over its use. We would also like to investigate the use of transfer learning to optimize a generically trained model, for multiple single households.

\bibliographystyle{plainnat}
\bibliography{library,custom_bib}

\end{document}

\endinput